\documentclass[11pt,letterpaper]{article}
\usepackage{naaclhlt2015}
\usepackage{times}
\usepackage{latexsym}
\usepackage{amsmath}
\usepackage{enumitem}
\usepackage{color}
\usepackage{url}

\title{Fast Rhetorical Structure Theory Discourse Parsing}

\author{Michael Heilman\\
        Educational Testing Service\\
        Princeton, NJ, USA \\
        {\tt mheilman@ets.org}
      \And
    Kenji Sagae\\
    Institute for Creative Technologies\\
    University of Southern California\\
    Los Angeles, CA, USA\\
  {\tt sagae@ict.usc.edu}}

\date{}

\begin{document}
\maketitle


\section{Introduction}

In recent years, There has been a variety of research on discourse parsing, particularly RST discourse parsing \cite{feng-hirst:2014:P14-1,li-EtAl:2014:P14-11,ji-eisenstein:2014:P14-1,joty-moschitti:2014:EMNLP2014,li-li-hovy:2014:EMNLP2014}.  Most of the recent work on RST parsing has focused on implementing new types of features or learning algorithms in order to improve accuracy, with relatively little focus on efficiency, robustness, or practical use.  Also, most implementations are not widely available.

Here, we describe an RST segmentation and parsing system that adapts models and feature sets from various previous work, as described below.  Its accuracy is near state-of-the-art, and it was developed to be fast, robust, and practical.  For example, it can process short documents such as news articles or essays in less than a second.

The system is written in Python and is publicly available at \url{https://github.com/EducationalTestingService/discourse-parsing}.

\section{Tasks and Data}
\label{sec:data}

We address two tasks in this work: discourse segmentation and discourse parsing.
Discourse segmentation is the task of taking a sequence of word and punctuation
tokens as input and identifying boundaries where new discourse units begin.
Discourse parsing is the task of taking a sequence of discourse units and
identifying relationships between them.  In our case, the set of these relationships
form a tree.

For both, we follow the conventions encoded in the RST Discourse Treebank \cite{rst-treebank}.
Here, we give a brief overview of the corpus.  See \newcite{Carlson:2001} for more information.

The treebank uses a representation where discourse is represented as a tree,
with labels on nodes indicating relationships between siblings.
Most RST relationships have a nucleus, expressing the core content,
and a satellite that contributes additional information to the nucleus.
Probably the simplest example is the ``attribution'' relationship:
attributed (e.g., quoted) text is labeled as the nucleus,
and text indicating the source of the attributed text is labeled
as the satellite, with an ``attribution'' subcategorization.

The leaves of the RST trees are ``elementary discourse units'' (EDUs),
which are contiguous spans of tokens roughly similar to indepedent clauses.
Most branching in RST trees is binary, with one satellite and one nucleus,
though there are some relations that have multiple nuclei and no satellite
(e.g., lists).

The RST corpus consists of a training set of 347 documents
and a test set of 38 documents.
The texts in the RST treebank are a subset of those in the Penn Treebank \cite{marcus:1993}.
For this reason, we retrained the syntactic parser used in our system, ZPar \cite{zhang-clark:2011},
on the subset 
of the Penn Treebank WSJ sections 2 to 21 texts
not present in the RST treebank.

For development of the system, we split the training set into a smaller
subset for model estimation and a development validation set
similar in size ($n = 40$) to the RST treebank test set.

\section{Discourse Segmenter Description}

In this section, we describe and evaluate the discourse segmentation component of the system.
Our discourse segmenter is essentially a reimplementation of the baseline system
from \newcite{xuanbach-leminh-shimazu:2012:SIGDIAL2012}.
We do not implement their reranked model, which is more complex to implement and probably less efficient,
and we use the ZPar parser \cite{zhang-clark:2011} for automatic syntactic parsing.

\subsection{Segmenter Model and Features}
\label{sec:segmentation}

Following \newcite{xuanbach-leminh-shimazu:2012:SIGDIAL2012}, we model RST as a tagging problem.
Specifically, for each token in a sentence, the system predicts whether that token is the
beginning of a new EDU or the continuation of an EDU.
For this task, we use a conditional random field \cite{lafferty2001conditional} model with $\ell_2$ regularization,
using the CRF++ implementation (\url{https://crfpp.googlecode.com}).
Also, we assume that a new sentence always starts a new EDU, regardless of the CRF output.

The CRF uses simple word and POS features as well as syntactic features.  The word and POS features are as follows (note that by ``word'', we mean word or punctuation token):

\begin{itemize}[itemsep=0em]
\item the lowercased form of the current word
\item the part-of-speech (POS) of the current word
\end{itemize}

The syntactic features are based on automatic parses from ZPar (using a retrained model as discussed in \S\ref{sec:data}).  For each of the following nodes in the syntactic tree,
there are two features, one for the nonterminal symbol and the head word (e.g., ``VP, said''),
and one for the nonterminal symbol and the head POS (e.g., ``VP, VBD'').  Note that these features will not be used for the last token in a sentence since there is no subsequent token.

\begin{itemize}[itemsep=0em]
\item $N_p$: the first common ancestor of the current token and the subsequent word
\item the subtree of $N_p$ that contains the current word
\item the subtree of $N_p$ that contains the subsequent word
\item the parent of $N_p$
\item the right sibling of $N_p$
\end{itemize}

All of these features are extracted for the current word, the previous 2 words, and next 2 words in the sentence.

\subsection{Segmenter Evaluation}

Following \newcite{xuanbach-leminh-shimazu:2012:SIGDIAL2012}, we evaluate segmentation performance using the gold standard EDUs from
the RST treebank test set, using the F1 score for the tag indicating
the beginning of a new EDU (``B-EDU'').
Since new sentences always begin new EDUs, we exclude the first tag in the output (always ``B-EDU'') for each sentence.
We first tuned the CRF regularization parameter using grid search
on the split of the training set used for development evaluations,
using a grid of powers of 2 ranging from 1/64 to 64.

The results are shown in Table~\ref{tab:segmentation}.
For comparison, we include previous results, including human-human agreement, reported by \newcite{xuanbach-leminh-shimazu:2012:SIGDIAL2012}, using syntax from the Stanford Parser \cite{klein-manning:2003:ACL} (it is not clear from the paper what parsing model was used).  The ``CRFSeg'' results are for the system from \newcite{Hernault:2010}.

We are uncertain as to the cause for the observed differences in performance, though we hypothesize that the differences are at least partially due to differences in syntactic parsing, which is a key step in feature computation.

\begin{table}[tb]
\begin{center}
\begin{tabular}{l|c|c|c}
& \textbf{P} & \textbf{R} & \textbf{F1}  \\ \hline

CRFSeg                     & 91.0 & 87.2 & 89.0  \\
Bach-etal-2012 (Base)      & 91.4 & 90.1 & 90.7  \\
Bach-etal-2012 (Reranking) & 91.5 & 90.4 & 91.0  \\
our system                 & 90.2 & 83.5 & 86.7  \\
Human agreement            & 98.5 & 98.2 & 98.3  \\

\end{tabular}
\caption{ Discourse segmentation performance in terms of percentages precision (``P''), recall (``R''), and F1 score (``F1'') for the ``B-EDU'' tag. \label{tab:segmentation}}
\end{center}
\end{table}

\section{Discourse Parser Description}

In this section, we describe our RST parser.
It borrows extensively from previous work, especially \newcite{sagae:2009:IWPT09}.\footnote{Note that we do not include \newcite{sagae:2009:IWPT09} in our evaluations since only within-sentence parsing performance was reported in that paper.}

\subsection{Shift-Reduce Approach}

Following \newcite{sagae:2009:IWPT09} and \newcite{ji-eisenstein:2014:P14-1}, we use an ``arc standard'' shift-reduce approach to RST discourse parsing.

\subsection{Parsing Model}

The parser maintains two primary data structures:
a queue containing the EDUs in the document that have not been processed yet, and a stack of RST subtrees that will eventually be combined to form a complete tree.

Initially, the stack is empty and all EDUs are placed in the queue.
Until a complete tree is found or no actions can be performed,
the parser iteratively chooses to perform shift or reduce actions.
The shift action creates a new subtree for the next EDU on the queue.

Reduce actions create new subtrees from the subtrees on the top of the stack.  There are multiple types reduce actions.  First, there are unary or binary versions of reduce actions, depending on whether the top 1 or 2 items on the stack will be included as children in the subtree to be created.  Second, there are versions for each of the nonterminal labels (e.g., ``satellite:attribution'').

Following previous work, we collapse the full set of RST relations to 18 labels.  Additionally, we binarize trees as described by \newcite{sagae-lavie:2005:IWPT}.

Following \newcite{sagae:2009:IWPT09} and \newcite{ji-eisenstein:2014:P14-1}, we treat the problem of selecting the best parsing action given the current parsing state (i.e., the stack and queue) as a classification problem.  We use multi-class logistic regression with an $\ell_1$ penalty, as implemented in the scikit-learn package, to estimate our classifier.

The parser supports beam search and $k$-best parsing, though we use simple greedy parsing (i.e., we set the beam size and $k$ to 1) for the experiments described here.

\subsection{Parsing Features}

To select the next shift or reduce action, the parsing model considers a variety of lexical, syntactic, and positional features adapted from various previous work on RST discourse parsing, such as that of \newcite{sagae:2009:IWPT09} and the systems we compare to in \S\ref{sec:experiments}.  The features are as follows:

\begin{itemize}[itemsep=0em]
\item the previous action (e.g., ``binary reduce to satellite:attribution'')
\item the nonterminal symbols of the $n$th subtree on the stack ($n = 0, 1, 2$), and their combinations
\item the nonterminal symbols of the children of the $n$th subtree on the stack ($n = 0, 1, 2$)
\item the lowercased words (and POS tags) for the tokens in the head EDU for the $n$th subtree on the stack ($n = 0, 1$) and the first EDU on the queue
\item whether, for pairs of the top 3 stack subtrees and the 1st queue item, the distance (in EDU indices) between the EDU head is greater than $n$ ($n = 1, 2, 3, 4$)
\item whether, for pairs of the top 3 stack subtrees and the 1st queue item, the head EDUs are in the same sentence
\item for the head EDUs of top 3 stack subtrees and the 1st queue item, the syntactic head word (lowercased), head POS, and the nonterminal symbol of the highest node in the subtree
\item syntactic dominance features between pairs of the top 3 stack items and 1st queue item, similar to \cite{Soricut:2003}
\item for each of the first 3 stack items or 1st queue item, whether that item starts a new paragraph
\end{itemize}

\section{Parsing Experiments}
\label{sec:experiments}


\begin{table*}[tb]
\begin{center}
\begin{tabular}{l|c|c|c|c}
& \textbf{syntax}
& \textbf{span}
& \textbf{nuclearity}
& \textbf{relation}  \\
\hline

our system                              & ZPar (retrained) & 83.5 & 68.1 & 55.1 \\
\newcite{li-li-hovy:2014:EMNLP2014}     & Stanford         & 84.0 & 70.8 & 58.6 \\
\newcite{joty-EtAl:2013:ACL2013}        & Charniak  (retrained)  & 82.5 & 68.4 & 55.7 \\
\newcite{joty-moschitti:2014:EMNLP2014} & Charniak  (retrained)  & --   & --   & 57.3 \\ 
\newcite{feng-hirst:2014:P14-1}         & Stanford         & 85.7 & 71.0 & 58.2 \\
\newcite{li-EtAl:2014:P14-11}           & Penn Treebank    & 82.9 & 73.0 & 60.6 \\
\newcite{ji-eisenstein:2014:P14-1}      & MALT             & 81.6 & 71.0 & 61.8 \\
\hline
Human agreement                         & --               & 88.7 & 77.7 & 65.8 \\

\end{tabular}
\caption{ Test set discourse parsing performance in terms of F1 scores (\%), using gold standard discourse segmentation.  ``syntax'' indicates the source of POS tags and syntactic parse trees:  
``Stanford'' refers to the Stanford parser \cite{klein-03}, ``MALT'' refers to \newcite{nivre-07}, and ``Charniak'' refers to \newcite{charniak-00}.
 \label{tab:test_set_results}}
\end{center}
\end{table*}


\begin{table*}[tb]
\begin{center}
\begin{tabular}{l|c|c|c|c}
& \textbf{syntax}
& \textbf{span}
& \textbf{nuclearity}
& \textbf{relation}  \\
\hline
our system                              & ZPar (retrained) & 83.5 & 69.3 & 57.4 \\
our system                              & PTB              & 84.7 & 71.2 & 59.4 \\
\hline

\end{tabular}
\caption{ Development set discourse parsing performance in terms of F1 scores (\%), using gold standard discourse segmentation.  ``syntax'' indicates the source of POS tags and syntactic parse trees. \label{tab:syntactic_parsing}}
\end{center}
\end{table*}


Following \cite[pp. 143--144]{marcu-book} and other recent work, we evaluate our system 
according to the F1 score over labeled and unlabeled spans of discourse units
in the RST treebank test set.  This evaluation is analogous to the \texttt{evalb} bracket
scoring program commonly used for constituency parsing (\url{http://nlp.cs.nyu.edu/evalb/}).
For comparison with previous results, we use gold standard discourse segmentations
(but automatic syntactic parses from ZPar).

We report F1 scores for agreement with the gold standard on unlabeled EDU spans (``span''), spans labeled only with nuclearity (``nuclearity''), and fully labeled spans that include relation information (``relation'').

We first tuned the $\ell_1$ regularization parameter using grid search
on the split of the training set used for development evaluations,
using a grid of powers of 2 ranging from 1/16 to 16.
We selected the setting that led to the highest F1 score for fully labeled spans (i.e., relation F1).

We compare to recently reported results from \newcite{ji-eisenstein:2014:P14-1} (their DPLP general +features model), \newcite{feng-hirst:2014:P14-1}, \newcite{li-EtAl:2014:P14-11}, \newcite{joty-moschitti:2014:EMNLP2014}, \newcite{li-li-hovy:2014:EMNLP2014}, and \newcite{joty-moschitti:2014:EMNLP2014}.\footnote{\newcite{joty-moschitti:2014:EMNLP2014} and \newcite{joty-moschitti:2014:EMNLP2014} do not explicitly state the source of syntactic parsers, but we infer from \newcite{joty-carenini-ng:2012:EMNLP-CoNLL} that the \newcite{charniak-00} parser was used, with a model trained on a subset of the Penn Treebank that did not include the RST treebank test set.}  The results are shown in Table~\ref{tab:test_set_results}.  The human agreement statistics were originally reported by \newcite{ji-eisenstein:2014:P14-1}.
For each system, the table indicates the source of POS tags and syntactic parse trees (``Penn Treebank'' means that gold standard Penn Treebank trees and tags were used).

We observe that our system is relatively close to the others in terms of F1 scores.
We hypothesize that the differences in performance are at least partially due to differences in syntactic parsing.

\subsection{The effect of automatic syntax parsing}

In order to show the effect of using automatic parsing, we report performance on the development set (\S\ref{sec:data}), using either gold standard syntax trees from the Penn Treebank or the automatic syntax trees from our retrained ZPar model (\S\ref{sec:data}) for computing features.  The F1 scores are shown in Table~\ref{tab:syntactic_parsing} (note that we are reporting results using the optimal settings from grid search on the development set).

It appears that the performance difference between using automatic rather than gold standard syntax is about 1 to 2 points of F1 score.

\subsection{Parsing Speed}

In this section, we evaluate the speed of the parser.  Most previous papers on RST parsing do not report runtime experiments, and most systems are not widely available or easy to replicate.

Our parser uses a shift-reduce parsing algorithm that has a worst-case runtime that is linear in the number of EDUs.  For comparison, \newcite{li-EtAl:2014:P14-11} employ a quadratic time maximum spanning tree parsing approach.  The approach from \newcite{joty-EtAl:2013:ACL2013} also uses a polynominal runtime algorithm.

Other linear time parsers have been developed \cite{feng-hirst:2014:P14-1,ji-eisenstein:2014:P14-1}.  However, feature computation can also be a performance bottleneck.  \newcite{feng-hirst:2014:P14-1} report an average parsing time of 10.71 seconds for RST treebank test set documents (and 5.52 seconds for a variant) on a system with ``four duo-core 3.0 GHz processors'', not including time for preprocessing or discourse segmentation.  In contrast, our system takes less than half a second per test set document on average (mean = 0.40, S.D. = 0.40, min. = 0.02, max. = 1.85 seconds) on a 2013 MacBook Pro with an i7-4850HQ CPU at 2.30 GHz.  Of course, these performance measurements are not completely comparable since they were run on different hardware.  The preprocessing (ZPar) and segmentation (\S\ref{sec:segmentation}) steps are also similarly fast.

\section{Conclusion}

In this paper, we have presented a fast shift-reduce RST discourse segmenter and parser.
The parser achieves near state-of-the-art accuracy and processes Penn Treebank documents in less than a second, which is about an order of magnitude faster than recent results reported by \newcite{feng-hirst:2014:P14-1}.

\section*{Acknowledgments}

We would like to thank Dan Blanchard, Xinhao Wang, and Keelan Evanini for feedback about the paper.
We would also like to thank Dan Blanchard, Diane Napolitano, Nitin Madnani, Aoife Cahill,
Chong Min Lee, Michael Flor, and Keisuke Sakaguchi for initial help with and feedback about
the implementation.

\bibliographystyle{naaclhlt2015}
\bibliography{nlp}

\end{document}